\documentclass[10pt,twocolumn,letterpaper]{article}

\usepackage{wacv}
\usepackage{times}
\usepackage{epsfig}
\usepackage{graphicx}
\usepackage{amsmath}
\usepackage{amssymb}
\usepackage{relsize}
\usepackage{url}
\usepackage{tabularx, booktabs}
\usepackage{pifont}
\usepackage{floatrow}

\usepackage[table]{xcolor}
\definecolor{lightgray}{gray}{0.9}
\definecolor{lightblue}{rgb}{0.93,0.95,1.0}
\definecolor{darkgreen}{rgb}{0.0,0.6,0.0}
\definecolor{mypink1}{rgb}{0.858, 0.188, 0.478}

\newcommand{\ignore}[1]{}

\newcommand\tripleq[3]{$\langle$\emph{#1}, \emph{#2}, \emph{#3}$\rangle$}

\newcolumntype{L}[1]{>{\raggedright\let\newline\\\arraybackslash\hspace{0pt}}m{#1}}
\newcolumntype{C}[1]{>{\centering\let\newline\\\arraybackslash\hspace{0pt}}m{#1}}
\newcolumntype{R}[1]{>{\raggedleft\let\newline\\\arraybackslash\hspace{0pt}}m{#1}}
\renewcommand{\eqref}[1]{Eq.~\ref{#1}}

\newcommand{\figref}[1]{Fig.~\ref{#1}}
\newcommand{\tabref}[1]{Tab.~\ref{#1}}
\newcommand{\secref}[1]{Sec.~\ref{#1}}

\newcommand{\reals}{\mathbb{R}}
\def\beq{\begin{equation}}
\def\eeq{\end{equation}}
\def\beqary{\begin{eqnarray}}
\def\eeqary{\end{eqnarray}}
\def\beqarz{\begin{eqnarray*}}
\def\eeqarz{\end{eqnarray*}}
\usepackage{boxedminipage}
\usepackage{bbold}

\renewcommand{\xi}{{\xx}^{(m)}}

\usepackage{xspace}

\newcommand{\SGs}{scene graphs\xspace}

\newcommand{\needcite}[1]{}

\newcommand{\be}{\begin{equation}}
\newcommand{\ee}{\end{equation}}
\newcommand{\benn}{\begin{equation*}}
\newcommand{\eenn}{\end{equation*}}
\newcommand{\bea}{\begin{eqnarray*}}
\newcommand{\eea}{\end{eqnarray*}}
\newcommand{\bean}{\begin{eqnarray}}
\newcommand{\eean}{\end{eqnarray}}

\newcommand{\xx}{\boldsymbol{x}} 
\renewcommand{\gg}{\boldsymbol{g}} 
\newcommand{\ff}{\boldsymbol{f}}

\newcommand{\zz}{\boldsymbol{z}}
\newcommand{\bv}{\boldsymbol{b}}
\newcommand{\rr}{\boldsymbol{r}}

\newcommand{\vv}{\boldsymbol{v}}

\newcommand{\comment}[1]{}

\newcommand{\polyring}[1]{\reals\left[x_1,\ldots,x_n\right]}

\definecolor{atomictangerine}{rgb}{0.8, 0.2, 0.1}
\definecolor{turq}{rgb}{0.0, 0.5, 0.5}
\definecolor{darkturq}{rgb}{0.0, 0.4, 0.4}
\definecolor{bright}{rgb}{0.8, 0.1, 0}
\definecolor{darkgray}{gray}{0.3}
\definecolor{mahogany}{rgb}{0.6, 0.05, 0.05}
\definecolor{pink}{rgb}{1,0.05,0.6}
\definecolor{myblue}{rgb}{0.3,0.05,0.9}

\renewcommand{\eqref}[1]{Eq.~\ref{#1}}

\newcommand{\nl}[1]{\emph{``#1"}}

\newcommand{\BBR}{Box Refiner}
\newcommand{\RRC}{Referring Relationships Classifier}

\wacvfinalcopy 

\def\wacvPaperID{336} 

\ifwacvfinal\fi
\begin{document}

\title{Differentiable Scene Graphs}

\author{
Moshiko Raboh$^{1^\star}$, \,\,
Roei Herzig$^{1^\star}$, \,\,
Jonathan Berant$^{1,4}$, \,\,
Gal Chechik$^{2, 3}$, \,\,
Amir Globerson$^{1}$ \vspace{3pt}\\
$^1$Tel Aviv University, $^2$Bar-Ilan University, $^3$NVIDIA Research, $^4$AI2 \\
}

\maketitle
\renewcommand*{\thefootnote}{$\star$}
\setcounter{footnote}{1}
\footnotetext{Equal Contribution.}
\renewcommand*{\thefootnote}{\arabic{footnote}}
\setcounter{footnote}{0}
\thispagestyle{empty}

\maketitle
\ifwacvfinal\thispagestyle{empty}\fi

\begin{abstract}
    Reasoning about complex visual scenes involves perception of entities and their relations. Scene Graphs (SGs) provide a natural representation for reasoning tasks, by assigning labels to both entities (nodes) and relations (edges). Reasoning systems based on SGs are typically trained in a two-step procedure: first, a model is trained to predict SGs from images, and next a separate model is trained to reason based on the predicted SGs. However, it would seem preferable to train such systems in an end-to-end manner. The challenge, which we address here is that scene-graph representations are non-differentiable and therefore it isn't clear how to use them as intermediate components. Here we propose \textit{Differentiable Scene Graphs} (DSGs), an image representation that is amenable to differentiable end-to-end optimization, and requires supervision only from the downstream tasks. DSGs provide a dense representation for all regions and pairs of regions, and do not spend modelling capacity on regions of the image that do not contain objects or relations of interest.  We evaluate our model on the challenging task of identifying referring relationships (RR) in three benchmark datasets: Visual Genome, VRD and CLEVR. Using DSGs as an intermediate representation leads to new state-of-the-art performance. The full code is available at \texttt{https://github.com/shikorab/DSG}.
\end{abstract}

\section{Introduction}
\label{sec:introduction}
Understanding the full semantics of rich visual scenes is a complex task that involves detecting individual entities, as well as reasoning about the combination of entities and the relations between them. To represent entities and their relations jointly, it is natural to view them as a graph, where nodes are entities and edges represent relations. Such representations are often called \textit{Scene Graphs} (SGs) \cite{johnson2015image}. Because SGs allow to explicitly reason about images, substantial efforts have been made to infer them from raw images \cite{img_retriev_using_sg, johnson2015image, sg_generation_msg_pass, support_relations, neural_motifs, mapping_to_imgs, InterpretableModel_nips18}.

\begin{figure}[t!]
    \begin{center}
    \includegraphics[trim=75 0 190 0, clip,width=\linewidth]{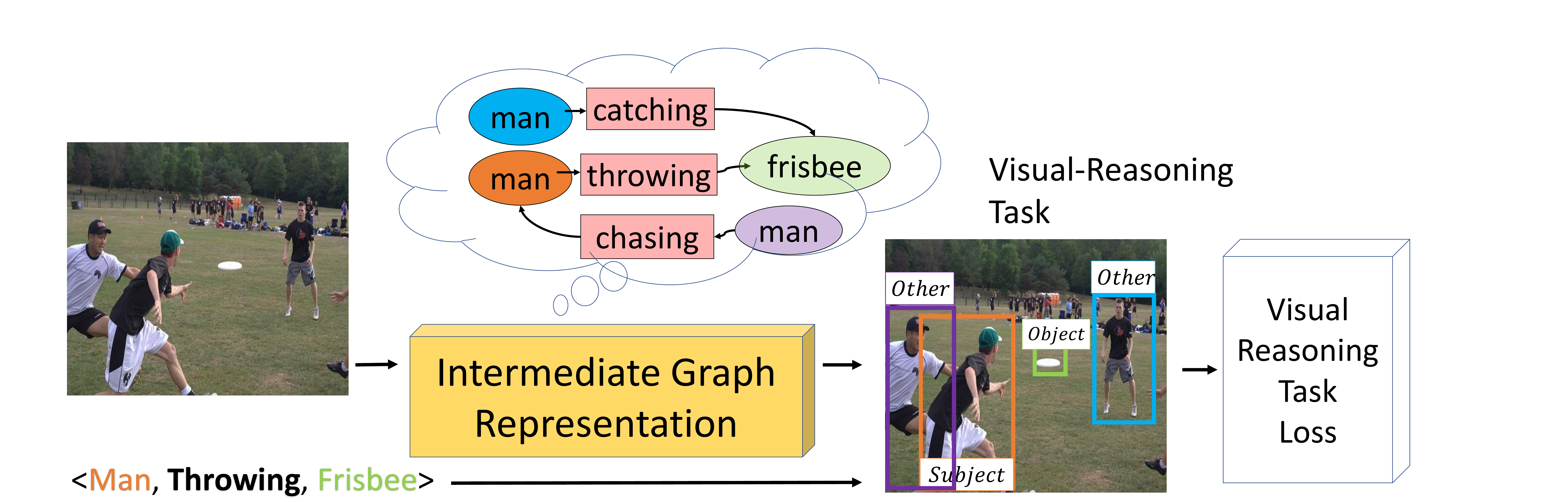}
    \vspace{-15pt}
    \caption{{\bf Differentiable Scene Graphs}: An intermediate ``graph-like" representation that provides a distributed representation for each entity and pair of entities in an image. Differentiable scene graphs can be learned with gradient descent in an end-to-end manner, only using supervision about a downstream visual reasoning task (referring relationships here).}
    \label{fig:teaser_fig}
    \end{center}
\end{figure}

While \SGs have been shown to be useful for various tasks \cite{img_retriev_using_sg, johnson2015image,johnson2018image}, using them as a component in a visual  reasoning system is challenging: (a)  Because \SGs are discrete representations, it is difficult to learn them in an end-to-end fashion from a downstream task. (b) The alternative is to pre-train SG predictors separately from supervised data, but this requires arduous and prohibitive manual annotation. Moreover, pre-trained SG predictors have \emph{low coverage}, because the set of labels they are pre-trained on rarely fits the needs of a downstream task. For example, given an image of a parade and a question \nl{point to the officer on the black horse}, that horse might not be a node in the graph, and the term ``officer" might not be in the vocabulary. Given these limitations, it is an open question how to make \SGs useful for visual reasoning applications.

In this work, we describe \textit{Differentiable Scene-Graphs} (DSG), which address the above challenges (Figure~\ref{fig:teaser_fig}). DSGs are an \textbf{intermediate representation trained end-to-end from the supervision for a downstream reasoning task}. The key idea is to relax the discrete properties of scene graphs such that each entity and relation is described with a dense differentiable descriptor. 

We demonstrate the benefits of DSGs in the task of resolving \emph{referring relationships} (RR) \cite{krishna2018referring} (see Figure 1). Here, given an image and a triplet query \tripleq{subject}{relation}{object}, a model has to find the bounding boxes of the subject and object that participate in the relation.

We train an RR model with DSGs as an intermediate component. As such, DSGs are not trained with direct supervision about entities and relations, but using several supervision signals about the downstream RR task. We evaluate our approach on three standard RR datasets: Visual Genome \cite{krishnavisualgenome}, VRD \cite{lang_prior} and CLEVR \cite{clevr}, and find that DSGs substantially improve performance compared to state-of-the-art approaches \cite{lang_prior,krishna2018referring}. 

To conclude, our novel contributions are: (1) A new \textit{Differentiable Scene-Graph} representation for visual reasoning, which captures information about multiple entities in an image and their relations. We describe how DSGs can be trained end-to-end with a downstream visual reasoning task without direct supervision of manually annotated scene-graphs. (2) A new architecture for the task of referring relationships, using DSGs as its central component. (3) New state-of-the-art results on the task of referring relationships on the Visual Genome, VRD and CLEVR datasets.

\begin{figure*}[t!]
    \begin{center}
    \includegraphics[width=\linewidth]{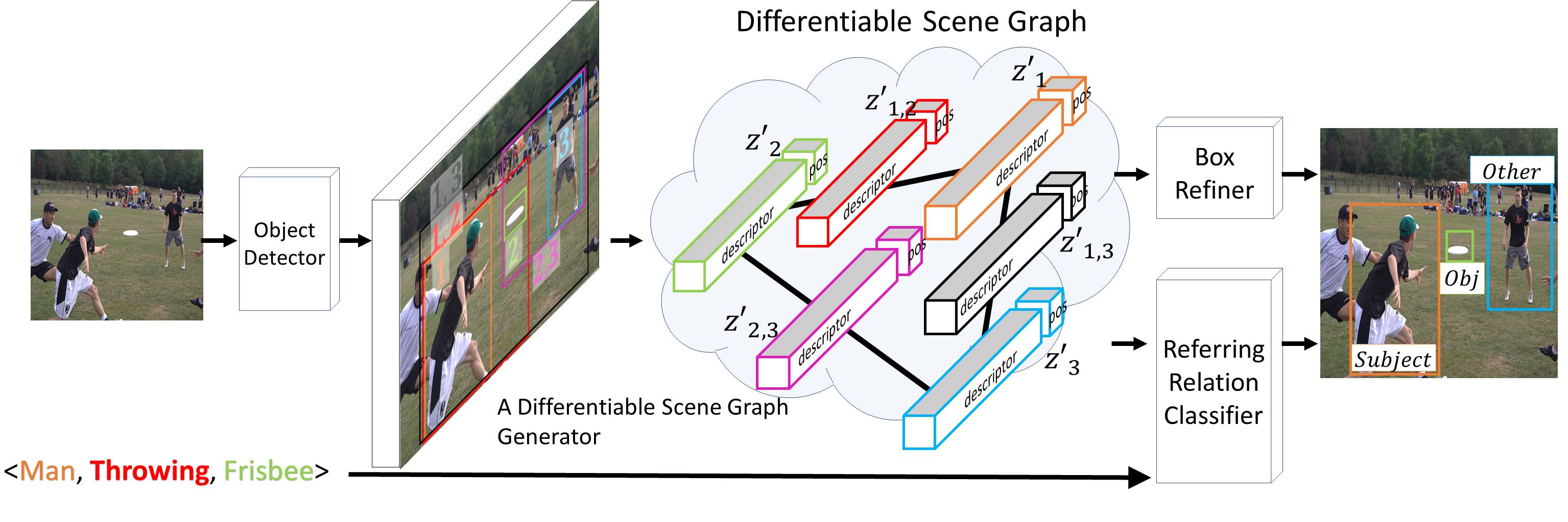}
        \vspace{-25pt}
        \caption{\textbf{The proposed architecture}. The input consists of an image and a relationship query triplet \tripleq{subject}{relation}{object}. (1) A detector produces a set of bounding box proposals. (2) An \textit{ROI-Align} layer extracts object features from the backbone using the boxes. In parallel, every pair of box proposals is used for computing a union box, and  pairwise features are extracted in the same way as object features. (3) These features are used as inputs to a Differentiable Scene-Graph Generator Module which outputs the Differential Scene Graph, a set of node and edge features that result from applying a graph convolutional network to the input features. (4) The DSG is used for both refining the original box proposals, as well as a Referring Relationships Classifier, which classifies each bounding box proposal as either \texttt{Subject}, \texttt{Object}, \texttt{Other} or \texttt{Background}. The ground-truth label of a proposal box will be \texttt{Other} if this proposal appears in another query relationship for this image. Otherwise the ground truth label will be \texttt{Background}. 
        }
        \vspace{-10pt}
        \label{fig:latent_graph}
    \end{center}
\end{figure*}

\section{Referring Relationship: The Learning Setup} 
\label{sec:setup}
In the referring relationship task \cite{krishna2018referring} we are given an image $I$ and a subject-relation-object query $q=\langle s,r,o \rangle$. The goal is to output a bounding box $\mathcal{B}_s$ for the subject, and another bounding box $\mathcal{B}_o$ for the object. In practice there are sometimes several boxes for each.
%
See \figref{fig:teaser_fig} for a sample query and expected output.

Following \cite{krishna2018referring}, we focus on training a referring relationship predictor from labeled data. Namely, we use a training set consisting of images, queries and the correct boxes for these queries. We denote these by $\{(I_j, q_j, (\mathcal{B}^s_j, \mathcal{B}^o_j)\}_{j=1}^N$. As in \cite{krishna2018referring}, we assume that the vocabulary of query components (subject, object and relation) is fixed. 

In our model, we break this task into two components that we optimize in parallel. We fine-tune the position of bounding boxes such that they cover entities tightly, and we also label each box as one of the following four possible labels. The labels ``Subject'' and ``Object'' correspond to the 's' and 'o' entities in the query. The label ``Other'' corresponds to boxes that contain entities (e.g., person or any other category that can appear as a subject or object in queries) that are not the subject or the object of the query. Finally, the label ``Background'' corresponds to boxes that do not contain an entity. We refer to the above two modules as \textbf{\BBR{}} and \textbf{\RRC{}}.

\section{Differentiable Scene Graphs}
\label{sec:dsgs}

We begin by discussing the motivation and potential advantages of using intermediate scene-graph-like representations, as compared to standard scene graphs. We then explain how DSGs fit into the full architecture of our model. 

\subsection{Why use intermediate DSG layers?}

A ``perfect'' scene graph (representing all entities and relations) captures most of the information needed for visual reasoning, and thus should be useful as an intermediate representation.
Such a SG can then be used by downstream reasoning algorithms, using the predicted SG as an input.  Unfortunately, learning to predict ``perfect'' scene graphs for any downstream task is unlikely. First, there is rarely enough data to train good SG predictors, and second, learning to predict SGs in a way that is independent of the downstream task, tends to yield less relevant SGs.  

Instead, we propose an intermediate representation, which we refer to as a ``Differentiable Scene Graph'' layer (DSG). A DSG captures the relational information as in a scene graph but can be trained end-to-end in a task-specific manner (\figref{fig:latent_graph}).
Like SGs, a DSG keeps descriptors for visual entities and their relations. Unlike SGs, whose nodes and edges are annotated by discrete values (labels), a DSG contains a dense distributed representation vector for each detected entity (referred to as a \emph{node descriptor}) and each pair of entities (referred to as an \emph{edge descriptor}). These representations are themselves learned functions of the input image (see supplementary for more details). Like SGs, a DSG only describes candidate boxes which cover entities of interest and their relations. 
Unlike SGs, each DSG descriptor captures not only the local information about a node, but also information about its context. 
Most importantly, because DSGs are differentiable, they are used as input to downstream visual-reasoning modules (in our case, a referring relationships module).

DGSs provide several computational and modelling advantages:

\noindent {\bf Differentiability.} Because node and edge descriptors are differentiable functions of detected boxes, and are fed into a differentiable reasoning module, the entire pipeline can be trained with gradient descent.

\noindent {\bf Dense descriptors.} By keeping dense descriptors for nodes and edges, the DSG keeps more information about possible semantics of nodes and edges, instead of committing too early to hard sparse representations. This allows it to better fit downstream tasks.

\noindent {\bf Supervision using downstream tasks.} Collecting supervised labels for training scene graphs is hard and costly. DGSs can be trained using training data that is available for downstream tasks, saving costly labeling efforts. On the other hand, when labeled scene graphs are available for given images, that data can be used when training the DSG, using an additional loss component.

\noindent {\bf Holistic representation.} DSG descriptors are computed by integrating global information from the entire image using graph neural networks (see supplemental materials). Combining information across the image increases the accuracy of object and relation descriptors.


\subsection{The DSG Model for Referring relationships}
\label{subsec:model_components}
We now describe how DSGs can be combined with other modules to solve a visual reasoning task. The architecture of the model is illustrated in  \figref{fig:latent_graph}. First, the model extracts bounding boxes for entities and relations in the image. Next, it creates a differentiable scene-graph over these bounding boxes. 
Then, DSG features are used by two output modules, aimed at answering a referring-relationship query: a \BBR{} module that refines the bounding box of the relevant entities, and a \RRC{} module that classifies each box as \texttt{Subject}, \texttt{Object}, \texttt{Other} or \texttt{Background}. We now describe these components in more detail.

\textbf{Object Detector.} We detect candidate entities using a standard region proposal network (RPN) \cite{faster_rcnn}, and denote their bounding boxes by $\bv{}_1,\ldots,\bv{}_B$ ($B$ may vary between images). We also extract a feature vector $\ff{}_i$ for each box and concatenate it with the box coordinates, yielding  $\zz_i = [\ff{}_i;\bv_i{}]$. See details in the supplemental material

\textbf{Relation Feature Extractor.} Given any two bounding boxes $\bv{}_i$ and $\bv{}_j$ we consider the smallest box that contains the two boxes (their ``union'' box). We denote this ``relation box'' by $\bv{}_{i,j}$ and its features by $\ff{}_{i,j}$. 
Finally, we denote the concatenation of the features $\ff{}_{i,j}$ and box coordinates $\bv_{i,j}{}$ by $\zz_{i,j} = [\ff{}_{i,j};\bv_{i,j}]$.

\begin{figure*}[t!]
\begin{center}
    \includegraphics[width=0.85\linewidth]{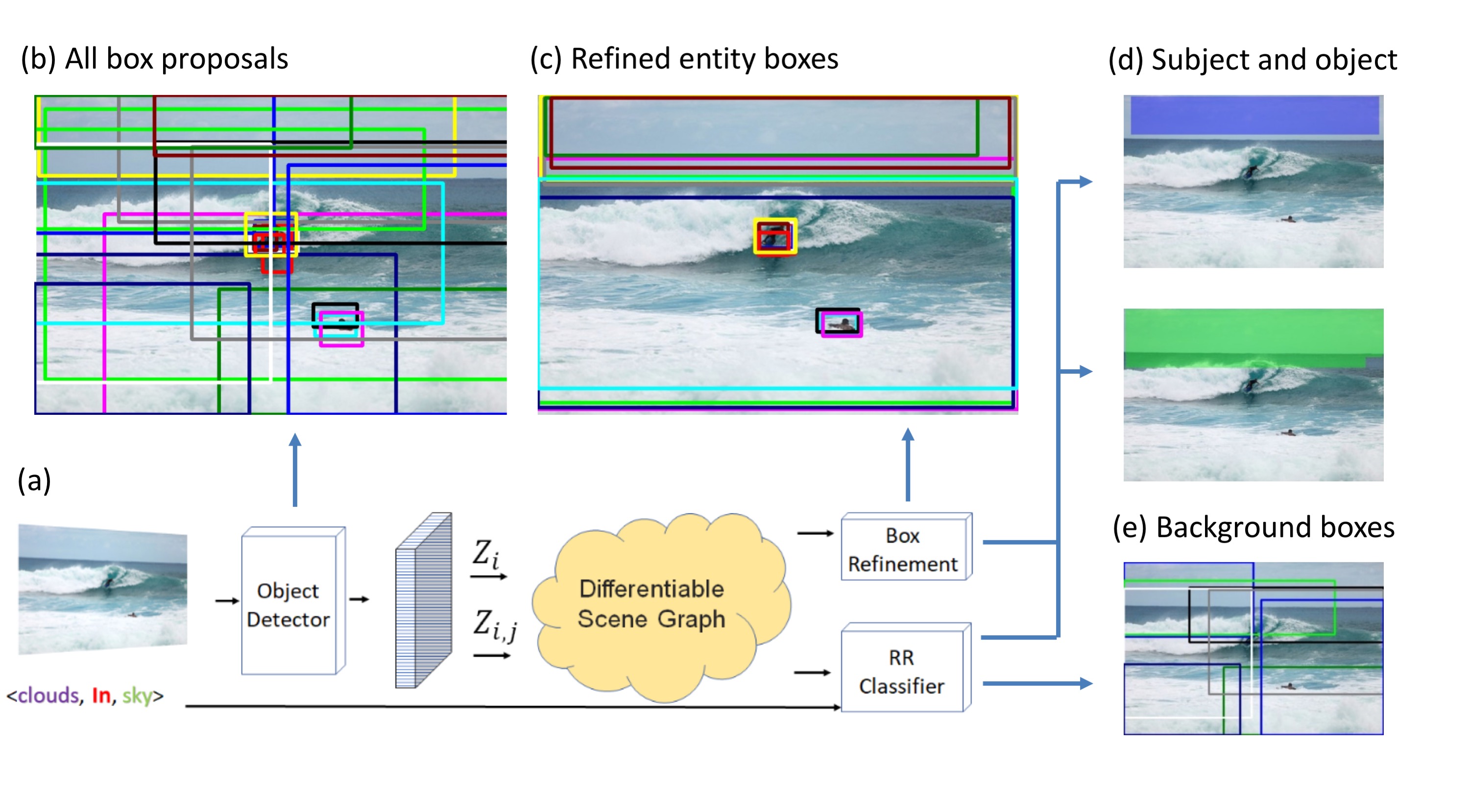}
    \vspace{-2em}
    \caption{The effect of box refinement and RR classification. (a) The DSG network is applied to an input image. (b) The \textit{object detector} component generates box proposals for entities in the image. (c) The \textit{RR classifier} component uses information from the DSG to label candidate boxes as \texttt{object} or \texttt{subject} entities. Then, the \textit{box refinement} component also uses DSG information, this time to improve box locations for those boxes labeled as entities by RR classifier. Here, boxes are tuned to focus on the most relevant entities in the image: the two ``men'', the ``surfboard'', the ``sky'' and the ``ocean''. (d) Once the RR classifier labels entity boxes, it can correctly refer to the entities in the query \tripleq{cloud}{in}{sky} (sky in green, clouds in violet). (e) Examples of candidate boxes classified as \texttt{background}.}
    \label{fig:boxes}
    \end{center}
    \vspace{-1em}
\end{figure*}

\textbf{Differentiable Scene-Graph Generator.} 
As discussed above, the goal of the DSG Generator is to transform the above features $\zz_i$ and $\zz_{i,j}$ into  differentiable representations of the underlying scene graph. Namely, map these features into a new set of dense vectors $\zz`_i$ and $\zz`_{i,j}$ representing entities and relations. 
This mapping is intended to incorporate the relevant context of each feature vector. Namely, the representation $z'_i$ contains information about the $i^{th}$ entity, together with its image-wide context.

There are various possible approaches to achieve this mapping. Here we use the model proposed by \cite{mapping_to_imgs}, which uses a graph neural network for this transformation (see supplemental material).

\hspace{10pt}
\textbf{Multi-task objective}. In many domains,  training with multi-task objectives can improve the accuracy of individual tasks, because auxiliary tasks operate as regularizers, pushing internal representations away from overfitting and towards capturing useful properties of the input. We follow this idea here and define a multi-task objective that has three components:
(a) a \RRC{} matches boxes to subject and object query terms. (b) A \BBR{} predicts accurate tight bounding boxes. (c) A Box Labeler recognizes visual entities in boxes if relevant ground truth is available.

\figref{fig:boxes} illustrates the effect of the first two components, and how they operate together to refine the bounding boxes and match them to the query terms. Specifically, \figref{fig:boxes}c, shows how box-refinement produces boxes that are tight around objects and subjects, and \figref{fig:boxes}d shows how RR classification matches boxes to query terms.

\textbf{(A) \RRC{}}. Given a DSG representation, we use it for answering referring relationship queries. Recall that the output of an RR query \tripleq{subject}{relation}{object} should be bounding boxes $\mathcal{B}_s, \mathcal{B}_o$ containing subjects and objects that participate in the query relation.
Our model has already computed $B$ bounding boxes $\bv{}_i$, as well as representations $\zz'_i$ for each box. We next use a prediction model $F_{RRC}(\zz'_i,q)$ that takes as input features describing the bounding box and the query, and outputs one of four labels $\{$\texttt{Subject}, \texttt{Object}, \texttt{Other}, \texttt{Background}$\}$ (see \secref{sec:setup}).
Denote the logits generated by this classifier for the $i^{th}$ box by $\rr_i\in\reals^4$. The output set $\mathcal{B}_s$ (or $\mathcal{B}_o$) is simply the set of bounding boxes classified as \texttt{Subject} (or \texttt{Object}). See supplemental materials for further implementation details.

\textbf{(B) \BBR{}}. The DSG is also used for further refinement of the bounding-boxes generated by the RPN network. The idea is that additional knowledge about image context can be used to improve the coordinates of a given entity. This is done via a network $F_{BR}(\bv_i, \zz'_i)$ that takes as input the RPN box coordinates and a differentiable representation $\zz'_i$ for box $i$, and outputs new bounding box coordinates. See \figref{fig:boxes} for an illustration of box refinement, and the supplemental material for further details.

\textbf{(C) Optional auxiliary losses: Scene-Graph Labeling}.
In addition to the \textit{\BBR{}} and \textit{\RRC{}} modules described above, one can also use supervision about labels of entities and relations if these are available at training time. 
Specifically, we train an object-recognition classifier operating on boxes, which predicts the label of every box for which a label is available. This classifier is trained as an auxiliary loss, in a multi-task fashion, and is described in detail below.


\section{Training with Multiple Losses}
\label{sec:train}
We next explain how our model is trained for the RR task, and how we can also use the RR training data for supervising the DSG component. We train with a weighted sum of three losses: 
(1) \RRC{} (2) \BBR{} (3) Optional Scene-Graph Labeling loss. 
We now describe each of these components. Additional details are provided in the supplemental material.


\subsection{Referring Relationship Classification Loss} 
\label{sec:train_RRC}
The \textit{\RRC{}} (\secref{subsec:model_components}) outputs logits $\rr_i$ for each box, corresponding to its prediction (\texttt{subject}, \texttt{object}, etc.). To train these logits, we need to extract their ground-truth values from the training data. Recall that a given image in the training data may have multiple queries, and so may have multiple boxes that have been tagged as subject or object for the corresponding queries. To obtain the ground-truth for box $i$ and query $q=\langle s,r,o \rangle$ we take the following steps. First, we find the ground-truth box that has maximal overlap with box $i$. If this box is either a subject or object for the query $q$, we set $\rr^{gt}_i$ to be \texttt{Subject} or \texttt{Object} respectively. Otherwise, if the overlap with a ground-truth box for a different image-query is greater than $0.5$, we set $\rr^{gt}_i=\texttt{Other}$  (since it means there is some other entity in the box), and we set $\rr^{gt}_i=\texttt{Background}$ if the overlap is less than $0.3$. If the overlap is in $[0.3,0.5]$ we do not use the box for training. For instance, given a query \tripleq{woman}{feeding}{giraffe} with ground-truth boxes for \nl{woman} and \nl{giraffe}, 
consider the box in the RPN that is closest to the ground-truth box for ``woman''. Assume the index of this box is $7$. Similarly, assume that the box closest to the ground-truth for ``giraffe' has index $5$. We would have $\rr_7^{gt}={\texttt{Subject}}$, $\rr_5^{gt}={\texttt{Object}}$ and the rest of the $\rr_i^{gt}$ values would be either \texttt{Other} or \texttt{Background}. Given these ground-truth values, the Referring Relationship Classifier Loss is simply the sum of cross entropies between the logits $\rr_i$ and the one-hot vectors corresponding to $\rr_i^{gt}$.

\subsection{\BBR{} Loss} 
\label{sec:train_BBR}
To train the \BBR{}, we use a smooth $L_1$ loss between the coordinates of the refined (predicted) boxes and their ground truth ones. 

\subsection{Scene-Graph Labeling Loss} 
\label{sec:train_SGL}
When ground-truth data about entity labels is available, we can use it as an additional source of supervision to train the DSG. Specifically, we train two classifiers. A classifier from features of entity boxes $\zz'_{i}$ to the set of entity labels, and a classifier from features  of relation boxes $\zz'_{i,j}$ to relation labels. We then add a loss to maximize the accuracy of these classifiers with respect to the ground truth box labels.


\ignore{Denote by $E$ the number of entity values (e.g., \nl{horse}, \nl{cat}, \nl{cow} etc.) and $R$ the number of relation values (e.g., \nl{holding}, \nl{kicking} etc.). We construct two classifiers. The first is a linear classifier that takes as inputs a vector $\zz'_i$ and outputs logits $\vv_i \in \reals^E$. The second is a linear classifier that takes as input a vector $\zz'_{i,j}$ output logits $\vv_{i,j} \in \reals^R$.  For each query, we have one triplet ground-truth for the above logits. Thus, our DSG Labeling Loss is a sum of cross-entropy loss for the three ground-truth labels (one for the subject, one for the object and one for the relation).
}

\subsection{Tuning the Object Detector} 
In addition to training the DSG and its downstream visual-reasoning predictors, the object detector RPN is also trained. 
The output of the RPN is a set of bounding boxes. The ground-truth contains boxes that are known to contain entities. The goal of this loss is to encourage the RPN to include these boxes as proposals. Concretely, we use a sum of two losses: First, an RPN classification loss, which is a cross entropy over RPN anchors where proposals of 0.8 overlap or higher with the ground truth boxes were considered as positive. Second, an RPN box regression loss which is a smooth L1 loss between the ground-truth boxes and proposal boxes.
%


\begin{figure*}[t!]
    \begin{center}
    \includegraphics[trim=0 0 770 0, clip,width=0.49\linewidth]{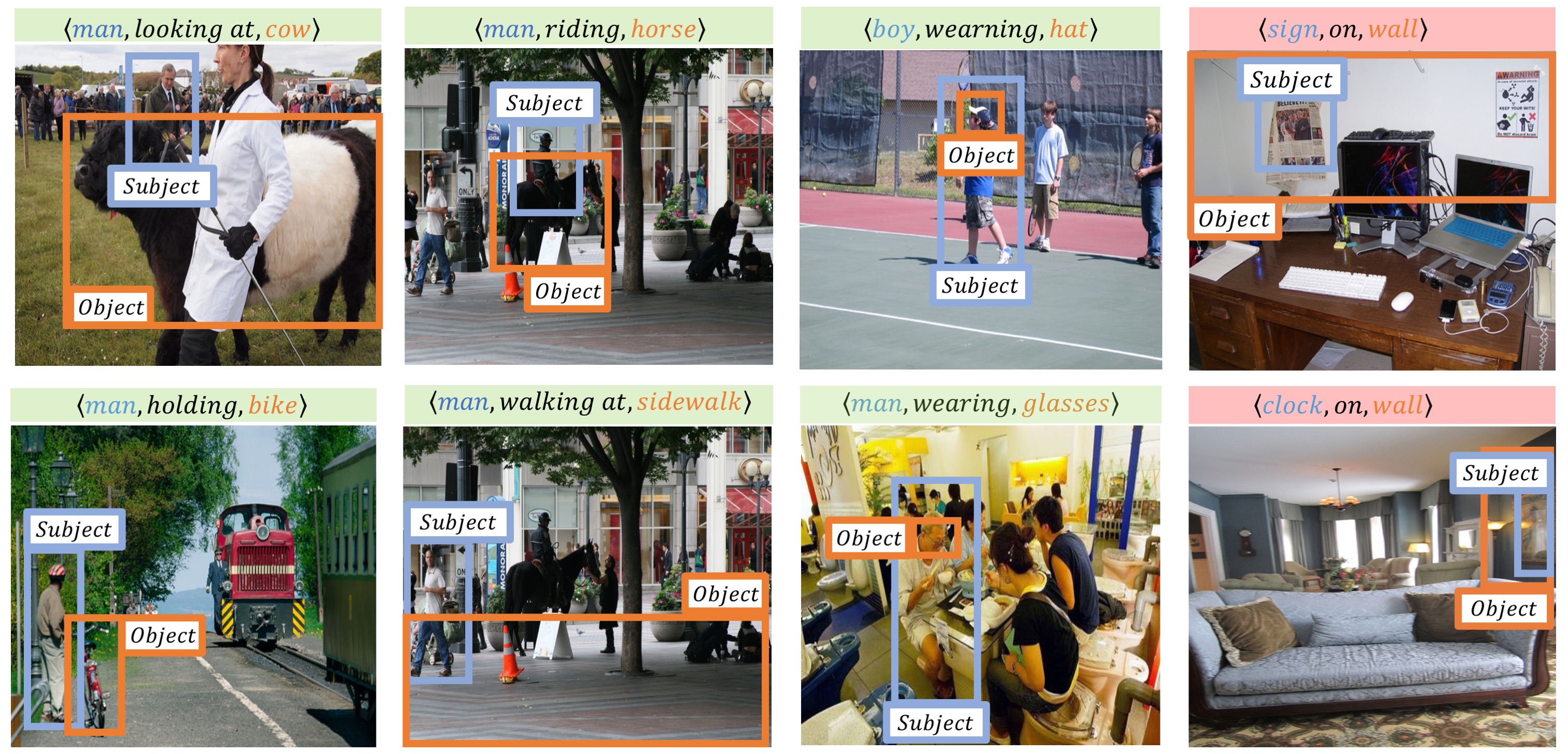}
    ~
    \includegraphics[width=0.49\linewidth]{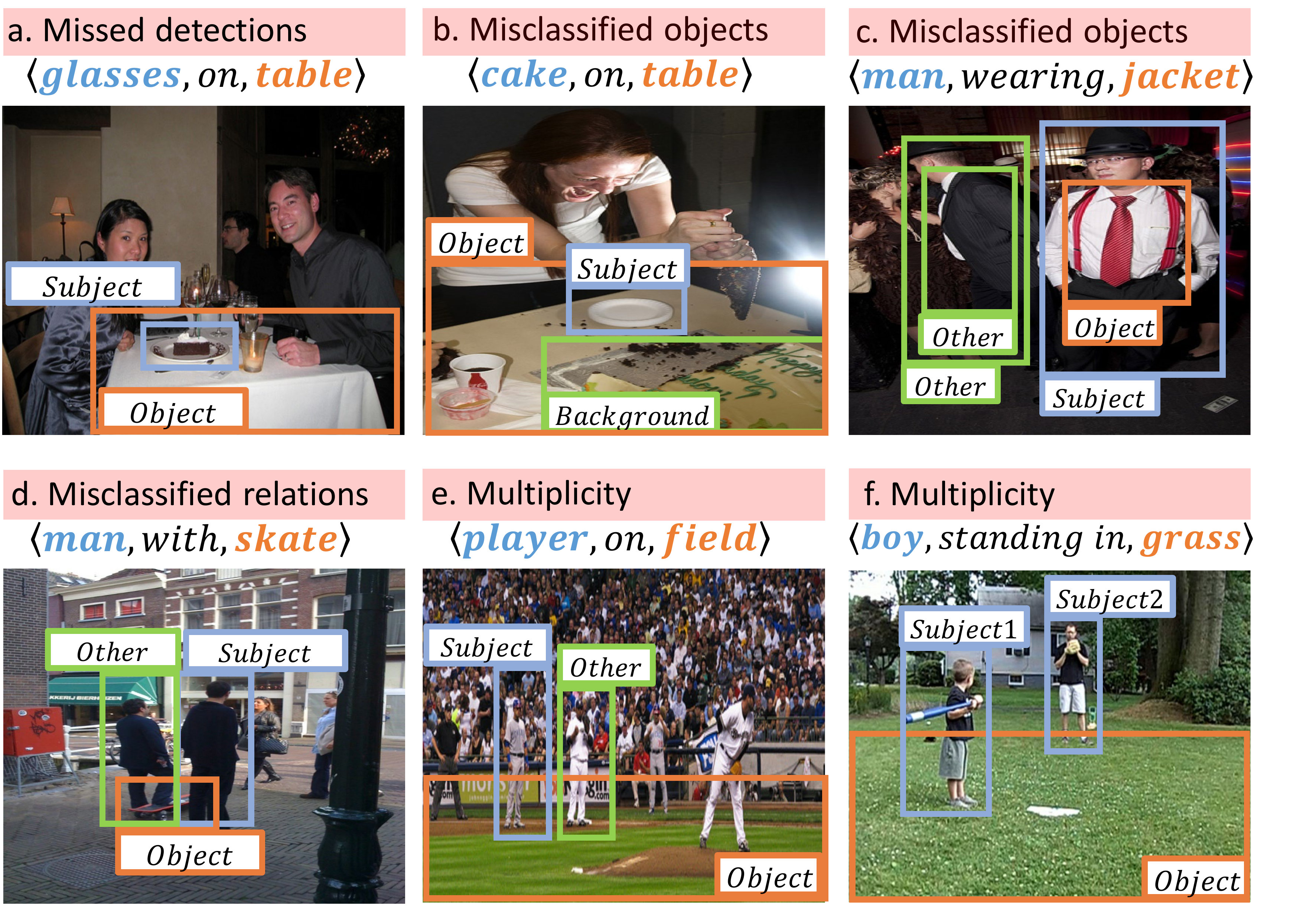}
        \vspace{-10pt}
        \caption{Qualitative examples demonstrating successful predictions of the DSG model (six left panels) and errors (six right panels).
        The right panels illustrate common failure cases for each error type. \textbf{a.} \textit{Missed detection:} the detector missed the glasses on the table. \textbf{b,c.} \textit{Misclassified object:}, the cake is detected but classified as a background. \textbf{d.} \textit{Misclassified relation:} The box classified as \textit{Subject} is indeed a man but it is not the man that has the required relation with the skate. \textbf{e,f.} \textit{Multiplicity}, Either too few or too many GT boxes are classified as \textit{Subject} or \textit{Object}.
        }
    \label{fig:win_examples}
    \end{center}
\end{figure*}

\begin{table}
    \small 
    \setlength\tabcolsep{4.5 pt}
    \begin{tabular}{lcccccc}
    \multicolumn{1}{c}{} & \multicolumn{6}{c}{Average IOU}\\
        \multicolumn{1}{c}{} & \multicolumn{2}{c}{Visual Genome} & \multicolumn{2}{c}{VRD} &   \multicolumn{2}{c}{CLEVR} \\
        & subject & object & subject & object & subject & object \\
        \midrule
        \scriptsize{\textsc{SS} \cite{shift}} & 0.399 & 0.469 & 0.320 & 0.371 & 0.740 & 0.740 \\
        \scriptsize{\textsc{CO}} \cite{cooccur2008} & 0.414 & 0.490 & 0.347 & 0.389 & 0.691 & 0.691 \\
        \scriptsize{\textsc{VRD} \cite{lang_prior}} & 0.417 & 0.480 & 0.345 & 0.387 & 0.734 & 0.732  \\
        \scriptsize{\textsc{SASS} \cite{krishna2018referring}} & 0.421 & 0.482 & 0.369 & 0.410 & 0.778 & 0.778\\
        \hline
        \scriptsize{\textsc{no-DSG}} & 0.412 & 0.47 & 0.333 & 0.366 & 0.937 & 0.937 \\
        \scriptsize{\textsc{DSG}}  & \textbf{0.489} & \textbf{0.539} & \textbf{0.4} & \textbf{0.435} & \textbf{0.963} & \textbf{0.963} \\
        \bottomrule
  \hline
  \end{tabular}
  \caption{\textbf{Comparison with baselines.} Test-set mean IOU in the referring relationship task for the baselines in \secref{sec:baselines} and the  Differentiable Scene Graph (DSG) model. Results are also reported for a \textsc{no-DSG} model (see \secref{sec:ablations}) which classifies the referring relationship directly from the RPN output.}
\hfill
\vspace{-1em}
\end{table}

\begin{table}
    \vspace{-1em}
    \begin{center}
    \begin{tabular}{lcc}
        \multicolumn{1}{l}{} & \multicolumn{2}{c}{Average IOU} \\
        & subject & object \\
        \midrule
        \scriptsize{\textsc{Two Step}} & $0.430 \pm  0.0014$ & $0.491 \pm 0.0014$ \\
        \scriptsize{\textsc{no-DSG}} & $0.405 \pm 0.0013$ & $0.461 \pm 0.0013$ \\
        \scriptsize{\textsc{DSG -SGL}} & $0.455 \pm 0.0014$ & $0.511 \pm 0.0013$ \\
        \scriptsize{\textsc{DSG -BR}} & $0.469 \pm  0.0014$ & $0.519 \pm 0.0014$ \\
        \scriptsize{\textsc{DSG}}  & $\textbf{0.477} \pm  0.0014$ & $\textbf{0.528} \pm 0.0014$\\
        \bottomrule
        \hline
        \end{tabular}
    \caption{{\bf Model ablations}: Results (including standard errors) for DSG variants on the validation set of the Visual Genome dataset. DSG values slightly differ from Table 1 which reports IOU on the test set. The various  models are described in \secref{sec:ablations}.}
    \label{results}
    \end{center}
    \vspace{-1em}
\end{table}

\section{Experiments}
\label{sec:experiments}
In the following sections we provide details about the  datasets, training, baselines models, evaluation metrics, model ablations and results. Implementation details of the model are provided in the supplemental material.

\subsection{Datasets}
\label{datasets}
We evaluate the model in the task of referring relationships on three datasets, each exhibiting a unique set of characteristics and challenges.
\newline
\textbf{CLEVR \cite{clevr}.} A synthetic dataset generated from scene-graphs with four spatial relations: ``left'', ``right'', ``front'' and ``behind'', and 48 entity categories. It has over 5M relationships where 33\% are ambiguous entities (multiple entities of the same type in an image).
\newline
\textbf{VRD \cite{lang_prior}.}  The Visual Relationship Detection dataset contains 5,000 images with 100 entity categories and 70 relation categories. In total, VRD contains 37,993 relationship annotations with 6,672 unique relationship types and 24.25 relations per entity category. 60.3\%  of these relationships refer to ambiguous entities.
\newline
\textbf{Visual Genome \cite{krishnavisualgenome}.} VG is the largest public corpus for visual relationships in real images, with 108,077 images annotated with bounding boxes, entities and relations. On average, images have 12 entities and 7 relations per image. In total, there are over 2.3M relationships where 61\% of those refer to ambiguous entities. 

For a proper comparison with previous results \cite{krishna2018referring}, we used the data from \cite{krishna2018referring} including the same entity and relation categories, query relationships and data splits.

\subsection{Evaluation Metrics}
We compare our model to previous work using the average IOU for subjects and for objects. To compute the average subject IOU, we first generate two $L \times L$ binary attention maps: one that includes all the ground truth boxes labeled as \texttt{Subject} (recall that few entities might be labeled as \texttt{Subject}) and the other includes all the box proposals predicted as \texttt{Subject}. If no box is predicted as \texttt{Subject}, the box with the highest score for the label \texttt{Subject} is included in the predicted attention map. We then compute the Intersection-Over-Union between the binary attention maps. For a proper comparison with previous work \cite{krishna2018referring}, we use $L=14$. The object boxes are evaluated similarly.

\subsection{Baselines}
\label{sec:baselines}
The Referring Relationship task was introduced recently \cite{krishna2018referring}, and the SSAS model was proposed as a possible approach (see below). We report the results for the baseline models in \cite{krishna2018referring}. 
When evaluating our Differentiable Scene-Graph model, we use exactly the evaluation setting as in \cite{krishna2018referring} (i.e.,  same  data splits, entity and relation categories). The baselines reported are: 
\ignore{
\textbf{(1) \textsc{Symmetric Stacked Attention Shifting (SSAS) \cite{krishna2018referring}}}: An iterative model that localizes the relationship entities using attention shift component learned for each relation. 

\textbf{(2) \textsc{Spatial Shifts \cite{shift}}}: Same as SSAS, but with no iterations and by replacing the shift attention mechanism with a simple statistical spatial shift for each relation label. 

\textbf{(3) \textsc{Co-Occurrence \cite{cooccur2008}}}: Uses an embedding of the subject and object pair for attending over the image features.

\textbf{(4) \textsc{Visual Relationship Detection (VRD) \cite{lang_prior}}}: Similar to Co-Occurrences model, but with an additional relationship embedding.
}
\begin{enumerate}
    \setlength{\itemsep}{0pt}%
    \setlength{\parskip}{0pt}%
    \item \textsc{Symmetric Stacked Attention Shifting (SSAS):} \cite{krishna2018referring} An iterative model that localizes the relationship entities using attention shift component learned for each relation. 
    \item \textsc{Spatial Shifts \cite{shift}}:  Same as SSAS, but with no iterations and by replacing the shift attention mechanism with a simple statistical spatial shift for each relation label. 
    \item \textsc{Co-Occurrence \cite{cooccur2008}}: Uses an embedding of the subject and object pair for attending over the image features.
    \item \textsc{Visual Relationship Detection (VRD) \cite{lang_prior}:} Similar to Co-Occurrences model, but with an additional relationship embedding.
\end{enumerate}

\section{Results}
Table 1 provides average IOU for \texttt{Subject} and \texttt{Object} over the three datasets described in \secref{datasets}. We compare our model to four baselines described in \secref{sec:baselines}. Our Differentiable Scene-Graph approach outperforms all baselines in terms of the average IOU.

\begin{figure}
    \begin{center}
        \includegraphics[width=0.8\linewidth]{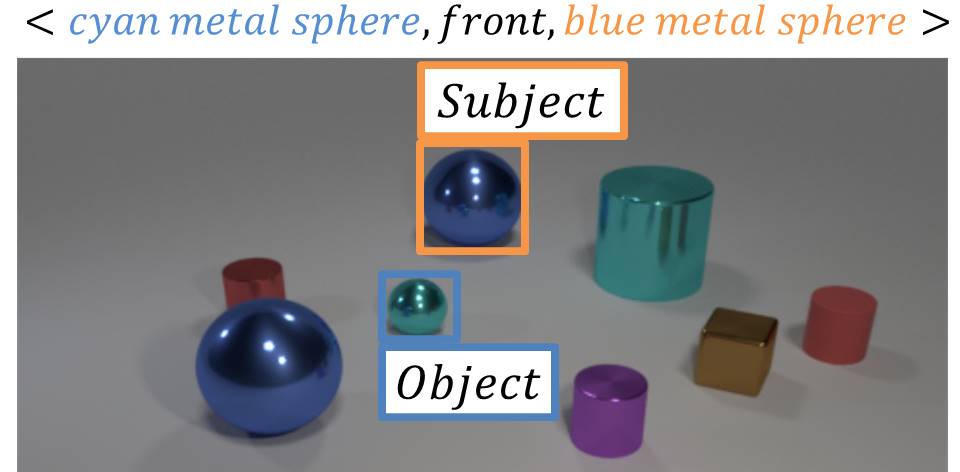}
        \caption{A typical image from the CLEVR \cite{clevr} dataset. The image was trimmed to focus on areas with visual content.} 
        \label{clevr_example}
    \end{center}
\end{figure}

Our results for the CLEVR dataset are significantly better than those in \cite{krishna2018referring}. Because CLEVR objects have a small set of distinct colors (Fig \ref{clevr_example}),  object detection in CLEVR is much easier than in natural images, making it easier to achieve high IOU. The baseline model without the DSG layer (\textsc{no-DSG}) is an end-to-end model with a two-stage detector in contrast to \cite{krishna2018referring} and already improves strongly over prior work with 93.7\%, and our novel DSG approach further improves to 96.3\% (reducing error by 50\%).

    

\subsection{Analysis of Success and Failure Cases.}
\figref{fig:win_examples} shows example of success cases and failure cases. We further analyzed the types of common mistakes and their distribution. Since DSGs depend on box proposals, they are sensitive to the quality of the object detectors. Manual inspection of images revealed four main error types: \textbf{(1)} 30\%: Detector failed:  the relevant box is missing from the box proposal list. \textbf{(2)} 23.3\% \textit{Subject} or \textit{Object} detected but classified as \textit{Other} or as \textit{Background}. \textbf{(3)} 16.6\%: Relation misclassified. The entities classified as \textit{Subject} and \textit{Object} match the query, but without the required relation. \textbf{(4)}  16.6\%: \emph{Multiplicity}. Either too few or too many of the GT boxes are classified as \textit{Subject} or \textit{Object}. \textbf{(5)} 13.3\%: Other, including incorrect GT, and hard-to-decide cases.


\subsection{Model Ablations}
\label{sec:ablations}
To gain further insight into the performance of the DSG model we performed the following ablations. First, since the model is trained with three loss components, we quantify the contribution of the Box Refinement loss and the Scene-Graph Labeling loss (it is not possible to omit the \RRC{} loss). We further evaluate the contribution of the DSG compared with a two-step approach which first predicts an SG, and then reasons over it. 
We compare the following models:
\ignore{
\textbf{(1) \textsc{DSG}}: The Differentiable Scene Graph model described in \secref{subsec:model_components} and trained as described in \secref{sec:train}.

\textbf{(2) \textsc{Two steps}}: A two-step model that first predicts a scene-graph, and then matches the query with the SG. The SG predictor consists of the same components used in the DSG: A box detector, DSG dense descriptors, and an SG labeler. It is trained with the same set of SG labels used for training the DSG. 

\textbf{(3) \textsc{DSG -SGL}}: DSG without the Scene-Graph Labeling component described in \secref{sec:train_SGL}). 

\textbf{(4) \textsc{DSG -BR}}: DSG where the \textit{\BBR{}} component of Section \ref{sec:train_BBR} is replaced with fine tuning the coordinates of the box proposal using the visual features $\ff^{}_i$ extracted by the Object Detector. 

\textbf{(5) \textsc{no-DSG}}: A baseline model that does not use the DSG representation. Instead, the model includes only an Object Detector and a RR classifier. The RR classifier uses the $\ff_i$ features extracted by the Object Detector instead of the $\zz’_i$ features. This model allows us to quantify the benefit of the differentiable scene representation for RR classification.
}
\begin{enumerate}
    \setlength{\itemsep}{2pt}%
    \setlength{\parskip}{2pt}%
     \item \textsc{DSG}: The Differentiable Scene-Graph model described in \secref{subsec:model_components} and trained as described in \secref{sec:train}.
     \item \textsc{Two steps}: Two-step model. We first predict a scene-graph, and then match the query with the SG. The SG predictor consists of the same components used in the DSG: A box detector, DSG dense descriptors, and an SG labeler. It is trained with the same set of SG labels used for training the DSG. Details in the supplemental material.
     \item \textsc{DSG -SGL}: DSG without the Scene-Graph Labeling component described in \secref{sec:train_SGL}). 
     \item \textsc{DSG -BR}: DSG where the \textit{\BBR{}} component of Section \ref{sec:train_BBR} is replaced with fine tuning the coordinates of the box proposal using the visual features $\ff^{}_i$ extracted by the Object Detector. This variant allows us to quantify the benefit of refining the box proposals based on the differentiable representation of the scene.
     \item  \textsc{no-DSG}: A baseline model that does not use the DSG representations. Instead, the model includes only an Object Detector and a RR classifier. The RR classifier uses the $\ff_i$ features extracted by the Object Detector instead of the $\zz’_i$ features. This model allows us to quantify the benefit of the differentiable scene representation for RR classification.
\end{enumerate}

Table 2 provides results of ablation experiments for the Visual Genome dataset \cite{krishnavisualgenome} on the validation set. 
All model variants based on scene representation perform better than the model that does not use the DSG representation (i.e., \textsc{DSG -SG}), demonstrating the power of contextualized scene representation. The \textsc{DSG} model outperforms all model ablations, illustrating the improvements achieved by using partial supervision for training the differentiable scene-graph. 
\figref{fig:ablation_examples} illustrates the effect of ablating various components of the model. 

\begin{figure}[t!]
    \begin{center}
    \includegraphics[width=0.99\linewidth]{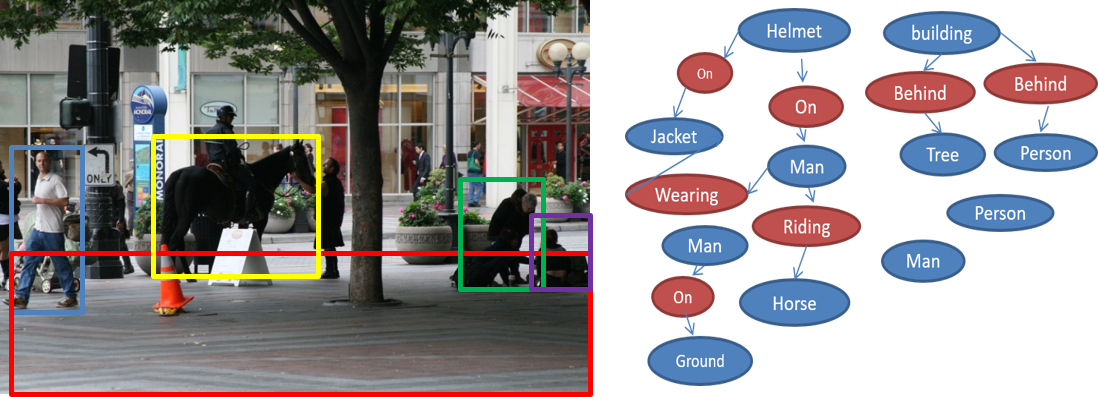}
    \caption{Inferring a Scene Graph from a DSG. Applying the RPN to this image results in 28 boxes. In (a) we show five of these, which received the largest weight in the attention model (details in the supplemental material.) within the DSG generator (\secref{subsec:model_components}). As mentioned in \secref{sec:train_SGL} in ``Scene-Graph Labeling Loss'' we can use the DSG for generating a labeled scene graph, corresponding to a fixed set of entities and relations. (b) shows this scene graph (i.e., the output of the classifiers predicting entity labels and relations), restricted to the largest confidence relations. It can be seen that most relations are correct, despite not having trained this model on complete scene graphs.}
    \label{fig:horse}
    \end{center}
\end{figure}

\subsection{Inferring SGs from DSGs}
The DSG is designed as a dense representation of objects and relations in the scene. It is thus natural to use it to predict these. This is easy to do in our context, since in \secref{sec:train_SGL} we in fact train such classifiers as an auxiliary task. Thus, for a given image we can construct a scene-graph out of the outputs of these objects and relation classifiers. 

\figref{fig:horse} illustrates the result of this process, showing a Scene-Graph inferred from the DSG. The predicted graph is indeed largely correct, even though it was not directly trained for this task (but rather from partial supervision). We further analyzed the accuracy of predicted SGs by comparing to ground-truth SG on visual genome (complete SGs were not used for training, only for analysis). SGs decoded from DSGs achieve accuracy of $76\%$ for object labels and  $70\%$ for relations (calculated for proposals with IOU $\ge 0.8$). 


\begin{figure}
    \begin{center}
        \includegraphics[width=\linewidth]{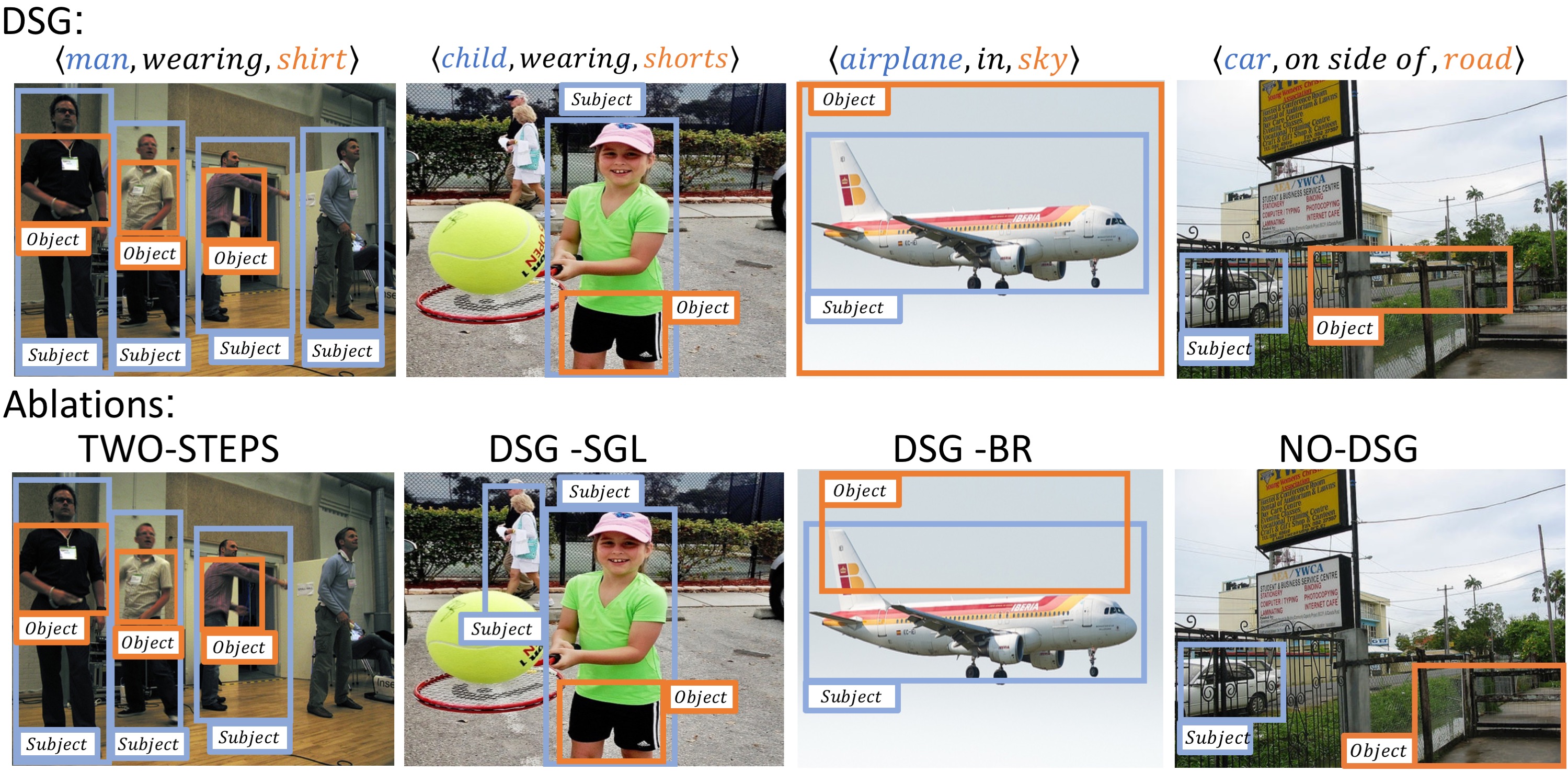}
        \caption{Comparing failures of ablations models with \textsc{DSG} predictions. The top row shows  \textsc{DSG} results, while the bottom row shows results from different ablations models as specified in ~\secref{sec:ablations}. In the first column, \textsc{Two Step} model the SG did not include the shirt of one of the the men, therefore this "subject" prediction was missed. In the second column, the \textsc{DSG -SGL} predicted failed to distinct between few entity classes `woman'' and ``child''. In the third column, the \textsc{DSG} refine the box of ``sky'' to cover all of the sky area. In the last column, the \textsc{NO-DSG} didn't classify the "object" box correctly.
        } 
        \label{fig:ablation_examples}
    \end{center}
    \vspace{-1em}
\end{figure}    

\section{Related Work}
\label{sec:related_work}
\textbf{Graph Neural Networks.} Recently, major progress has been made in constructing graph neural networks (GNN). These refer to a class of neural networks that operate directly on graph-structured data by passing local messages \cite{gilmer2017neural, Li2015GatedGS}. Variants of GNNs have been shown to be highly effective at relational reasoning tasks \cite{nn_for_relational}, classification of graphs \cite{bruna2013invariant, dai_disc_embs, niepert16, DefferrardBV16} and classification of nodes in large graphs \cite{graph_conv, inductive_repr_large_grphs}. The expressive power of GNNs has also been studied in \cite{mapping_to_imgs, deep_sets}. GNNs have also been applied to visual understanding in \cite{mapping_to_imgs, graph_rcnn, Wang_videogcnECCV2018, herzig2019STAG} and control \cite{Gonzalez18, relational_rl_deepmind18}. Similar aggregation schemes have also been applied to object detection \cite{hu2018relation}. Our goal here is to generate DSG such that each object descriptor encompasses not only the local information about the object, but also information about its context within the scene. To achieve this we use the GNN proposed by \cite{mapping_to_imgs}.

\textbf{Visual Relationships and Scene Graphs.} Earlier work aimed to leverage visual relationships for improving  detection \cite{VisualPhrases}, action recognition \cite{herzig2019STAG}, few shot \cite{chen2019scene, dornadula2019visual}, pose estimation \cite{DesaiR12}, semantic image segmentation \cite{GuptaD08} or detection of human-object interactions  \cite{yang2017support, plummerPLCLC2017, Li_2019_CVPR}. Lu {\em et al.} \cite{lang_prior} were the first to formulate detection of visual relationships as a separate task. They learn a likelihood function that uses a language prior based on word embeddings for scoring visual relationships and constructing SGs. SGs provide a compact representation of the semantics of an image. Previous SG prediction works used attention \cite{Qi_2019_CVPR, Gkanatsios_2019_ICCV} or neural message passing \cite{sg_generation_msg_pass}. \cite{pixels_to_graph} suggested to predict graphs directly from pixels in an end-to-end manner. \cite{neural_motifs} considers global context using an RNN by reading sequentially the independent predictions for each entity and relation and then refines those predictions.
SGs have been shown to be useful for semantic-level interpretation and reasoning about a visual scene \cite{johnson2018image, ashual2019specifying, herzig2019canonical, Schroeder_2019_ICCV}. Extracting SGs from images provides a semantic representation that can later be used for reasoning, question answering \cite{yi2018neural, hu2019language, Liang_2019_ICCV}, and image retrieval \cite{img_retriev_using_sg, entangled_scene}. Using SGs for reasoning tasks is challenging. Instead, we propose an intermediate representation which captures the relational information as in SGs but can be trained end-to-end in a task-specific manner.

\textbf{Referring Relationships.} The RR task is closely related to the task of referring expressions, where an entity in an image needs to be identified given a natural language expression. Several recent works considered using context for this task \cite{kazemzadeh2014referitgame, krahmer2012computational, Chen_2019_ICCV, Tanaka_2019_ICCV, Yang_2019_ICCV, Liu_2019_ICCV}. \cite{mao2016generation} described a model that has two parts: one for generating expressions that point to an entity in a discriminative fashion and a second for understanding these expressions and detecting the referred entity. \cite{yu2016modeling} explored the role of context and visual comparison with other entities in referring expressions. Modelling context was also the focus of \cite{nagaraja2016modeling}, using a multi-instance-learning objective. RR \cite{krishna2018referring} as opposed to referring expression, focuses on the vision side rather than the language side by forming a simple structured query that requires modeling interactions between the image entities. \cite{krishna2018referring} also introduce an explicit iterative model that localizes the two entities in the RR task, conditioned on one another. We use the RR task to demonstrate the power of our semantic latent representation, resulting in a new state of the art results on three vision datasets that contains visual relationships.

\section{Conclusion}
\label{sec:conclusion}
This work is motivated by the assumption that accurate reasoning about images may require access to a detailed representation of the image. While scene graphs provide a natural structure for representing relational information, it is hard to train very dense SGs in a fully supervised manner, and for any given image, the resulting SGs may not be appropriate for downstream reasoning tasks. Here we advocate DSGs, an alternative representation that captures the information in SGs, which is continuous and can be trained jointly with downstream tasks. Our results, both qualitative (Fig \ref{fig:win_examples} ) and quantitative (Table 1,2), suggest that DSGs effectively capture scene structure, and that this can be used for down-stream tasks such as referring relationships. 

One natural next step is to study such representations in additional downstream tasks that require integrating information across the image. Some examples are caption generation and visual question answering. DSGs can be particularly useful for VQA, since many questions are easily answerable by scene graphs (e.g., counting questions and questions about relations). Another important extension to DSGs would be a model that captures high-order interactions, as in a hyper-graph. Finally, it will be interesting to explore other approaches to training the DSG, and in particular finding ways for using unlabeled data for this task.
\newline
\newline
{\bf Acknowledgments:} This project was funded by the European Research Council (ERC) under the European Unions Horizon 2020 research and innovation programme (grant ERC HOLI 819080).
{\small
    \bibliographystyle{ieee}
    \bibliography{rr}
}

\newpage
\setcounter{section}{0} 

\section*{Supplementary Material} 

This supplementary material includes: (1) Model implementation details. (2) Details about the reasoning component in two steps ablation module.

\section{Model Details}
\label{apendix:Supplementary}

The model in Sec. 3.2 is implemented as follows.

\textbf{Object Detector and Relation Feature Extractor.} For object detection, we used Faster-RCNN with a 101-layers ResNet backbone. The RPN was trained with anchor scales of $\{4, 8, 16, 32\}$ and aspect ratios $\{0.5, 1, 2\}$. RPN proposals were filtered by non-maximum suppression with IOU-threshold of 0.5 and score higher than 0.8. We use at most 32 proposals per image. Both the entity features $\ff^{}_i$ and the relation features $\ff^{}_{i,j}$ are first extracted from the convolutional network feature map by the ROI-Align layer as $7 \times 7 \times 2048$ features. They are then reduced to a $7 \times 7 \times 512$ by convolution layer of size $1 \times 1$ and finally reduced to $1 \times 512$ by an average pooling layer.

\textbf{Referring Relationship Classifier.} The referring relationship classifier $F_{RRC}$ is a fully-connected network with two layers of 512 hidden units each.

\textbf{Bounding Box Refinement.} The box refinement model applies a linear function to $\zz_i$ to obtain four outputs $[dx, dy, dw, dh]$.
Denote the RPN box by $ [x, y, w, h]$. The refined box is then: $[dx \cdot w + x, dy \cdot h + y, e^{dw} \cdot w, e^{dh} \cdot h]$ (as in the correction used by Faster-RCNN).

\textbf{Computational Estimation.} Our model creates a graph with $n$ nodes for objects and $n^2$ edges for relations. In the datasets we analyzed, using  $n=32$ objects within an image is sufficient. Adding the DSG component has a limited effect on complexity and run time. Specifically, as shown in \tabref{analysis}, the DSG component adds 4M parameters and up to 1.5G operations (when  $n=32$) which is \textbf{only 10\% of the parameters} and number of operations of the backbone network ($\sim$40M parameters and $\sim$15G operations). Adding DSG \textbf{increases training time by only 15\%}. This is largely thanks to the fact that all $n^2$ relations can parallelized.

\begin{table}[h]
\vspace{-1em}
\begin{center}
  \begin{tabular}{L{3.1CM}C{2.4CM}C{1.5CM}}
  \hline
  & DSG Generator & Resnet101 \\
        \midrule
        Trainable parameters & $< 4M$ & $> 40M$ \\
        Number of operations & $< 1.5G$ & $> 15G$\\
        \bottomrule
  \hline
  \end{tabular}
  \begin{tabular}{L{3.1CM}C{2.4CM}C{1.5CM}}
  \hline
  & \textsc{DSG} &  \textsc{DSG -SG}\\
        \midrule
        Running time [sec] & 0.054 & 0.045 \\
        Training time  [sec] & 0.19 & 0.165 \\
        \bottomrule
  \hline
  \end{tabular}
  \caption{Analysis of running/training time and computational resources of DSGs.}
  \label{analysis}
\end{center}
\vspace{-1em}
\end{table}

\textbf{Differentiable Scene Graph Generator.} We next describe the module that takes as input features $\zz_i$ and $\zz_{i,j}$ extracted by the RPN and outputs a set of vectors $\zz'_i,\zz'_{ij}$ corresponding to Differentiable Scene-Graph over entities and relationships in the image. For this model, we use the Graph Permutation Invariant (GPI) architecture introduced in \cite{mapping_to_imgs}. A key property of this architecture is that it is invariant to permutations of the input that do not affect the labels.

The GPI transformation is defined as follows. First, the set of all input features is summarized via a permutation-invariant transformation into a single vector $\gg$:
\begin{equation}\label{sgp_eq}
\gg = \sum_{i=1}^n \boldsymbol{\alpha} (\zz_i, \sum_{j \neq i} \boldsymbol{\phi}(\zz_{i}, \zz_{i,j}, \zz_{j}))
\end{equation}
Here $\boldsymbol{\alpha}$ and $\boldsymbol{\phi}$ are fully connected networks. Then the new representations for entities and relations are computed via:
\begin{equation}
    \zz'_{k} = {\boldsymbol\rho}^{entity}(\zz_{k},\gg) \ , \
    \zz'_{k,l} = {\boldsymbol\rho}^{relation}(\zz_{k,l},\gg)     
\end{equation} 
where ${\boldsymbol{\rho}}$ above are fully connected networks. 

The three networks $\boldsymbol{\phi}$, $\boldsymbol{\alpha}$ and $\boldsymbol{\rho}$, described in GPI architecture are two fully-connected layers with 512 hidden units. The output size of $\phi$ and $\alpha$ is 512, and of $\boldsymbol{\rho}$ is 1024. We used the version with integrated attention mechanism replacing the sum operations in equation \ref{sgp_eq}.

\section{Model Ablations}
Additional details about the \textsc{Two Step} model:
Recall that in Two-step model a scene graph is first predicted, followed by a reasoning module. The reasoning module gets as an input a query \tripleq{subject}{relation}{object} and a scene graph and outputs the nodes that represents the subject and the nodes that represents the object. In case the triplet \tripleq{subject}{relation}{object} exists in the scene graph, the reasoning module simply returns the involved nodes. Otherwise, it selects the triplet in the scene graph that has the highest probability to be the required triplet according to the probabilities provided by the scene graph.

\end{document}